\pdfoutput=1

\documentclass[11pt]{article}

\usepackage[final]{acl}
\usepackage{comment}
\usepackage{times}
\usepackage{latexsym}
\usepackage{amsmath} 

\usepackage[T1]{fontenc}

\usepackage[utf8]{inputenc}

\usepackage{microtype}

\usepackage{inconsolata}

\usepackage{graphicx}

\title{MiMIC: Multi-Modal Indian Earnings Calls Dataset to Predict Stock Prices}

\author{Sohom Ghosh \\
  Jadavpur University \\
  Kolkata, India \\
  \texttt{sohom1ghosh@gmail.com} \\\And
  Arnab Maji \\
  Independent Researcher \\
  Kolkata, India \\
  \texttt{arnabmaji09@gmail.com} \\ \\\And
  Sudip Kumar Naskar \\
  Jadavpur University \\
  Kolkata, India \\
  \texttt{sudip.naskar@gmail.com} \\}

\begin{document}
\maketitle
\begin{abstract}
Predicting stock market prices following corporate earnings calls remains a significant challenge for investors and researchers alike, requiring innovative approaches that can process diverse information sources. This study investigates the impact of corporate earnings calls on stock prices by introducing a multi-modal predictive model. We leverage textual data from earnings call transcripts, along with images and tables from accompanying presentations, to forecast stock price movements on the trading day immediately following these calls. To facilitate this research, we developed the \textbf{MiMIC} (\textbf{M}ult\textbf{i}-\textbf{M}odal \textbf{I}ndian Earnings \textbf{C}alls) dataset, encompassing companies representing the Nifty 50, Nifty MidCap 50, and Nifty Small 50 indices. The dataset includes earnings call transcripts, presentations, fundamentals, technical indicators, and subsequent stock prices. We present a multimodal analytical framework that integrates quantitative variables with predictive signals derived from textual and visual modalities, thereby enabling a holistic approach to feature representation and analysis. 
This multi-modal approach demonstrates the potential for integrating diverse information sources to enhance financial forecasting accuracy. To promote further research in computational economics, we have made the MiMIC dataset publicly available under the CC-NC-SA-4.0 licence. Our work contributes to the growing body of literature on market reactions to corporate communications and highlights the efficacy of multi-modal machine learning techniques in financial analysis.
\end{abstract}

\section{Introduction}

\begin{figure}[t]
  \includegraphics[width=\columnwidth]{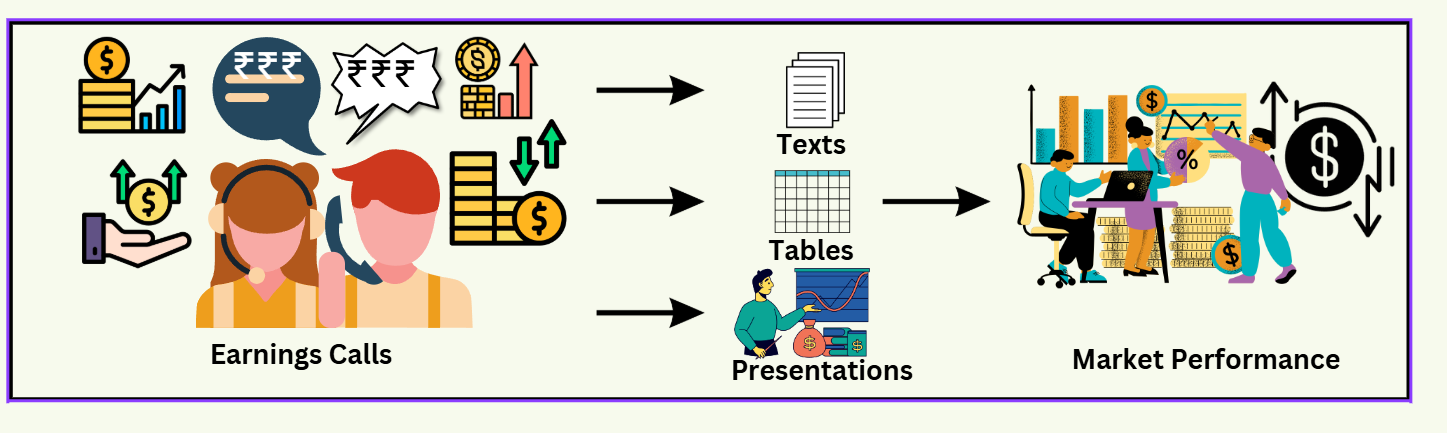}
  \caption{Multi-modal analysis of Earning Calls of Indian Companies}
  \label{fig:multi-earnings-intro}
\end{figure}

In financial markets, earnings call presentations serve as critical sources of forward-looking information, integrating verbal discourse (text transcripts), visual aids (charts, diagrams), and quantitative data (tables) to communicate corporate performance and strategic outlook. While existing research has explored the predictive power of textual, audio, and numerical data from earnings calls in forecasting stock price movements, the synergistic integration of multi-modal data — text, visual, and tabular — remains underexplored. Current approaches often discard the visual representations, potentially overlooking nuanced interactions between qualitative narratives, visual representations of financial metrics, and structured tabular data. Moreover, there is a notable dearth of research focusing on earnings calls within the Indian context, highlighting a significant gap in understanding how these events impact stock market dynamics in this specific domain.


This study aims to advance financial forecasting methodologies by proposing a novel architecture that bridges modality-specific representations while addressing the inherent complexity of real-world earnings communication. We present this in Figure \ref{fig:multi-earnings-intro}. 
The outcomes of this study seek to enhance decision-making for investors, analysts, and automated trading systems reliant on timely interpretation of multi-modal financial disclosures.

\textbf{Our contributions are:}\\
$\bullet$ \textbf{MiMIC Dataset:} We introduce MiMIC (\textbf{M}ult\textbf{i}-\textbf{M}odal \textbf{I}ndian Earnings \textbf{C}alls), a novel dataset specifically curated for analyzing Indian financial markets. MiMIC comprises earnings call transcripts and accompanying presentations from Indian companies, coupled with their corresponding stock market performance data on the day following the release of quarterly results. To the best of our knowledge, this is the first multi-modal dataset of this nature for the Indian market.\\
$\bullet$ \textbf{Multi-Modal Predictive Framework:} We present a comprehensive multi-modal framework designed to predict stock prices. This framework integrates information from earnings calls, company fundamentals, technical indicators, and broader market variables to provide a holistic view of factors influencing stock performance.

\section{Related Work}
The analysis of earnings calls for stock price prediction has become a prominent area in financial research, driven by advancements in multi-modal data integration, including text, images, and tables. Earnings calls serve as a vital information repository, offering insights that extend beyond conventional financial indicators. Research by Medya et al. \cite{medya-www-22} demonstrates the predictive power of semantic elements within earnings call transcripts. Their findings indicate that the narrative structure and tonal qualities of corporate communications during these calls substantially shape investor sentiment and consequent market reactions. Huynh and Shenai \cite{huynh2019option} document an inverse relationship between option trading volumes and immediate stock price reactions following earnings announcements.

A significant breakthrough came with the development of models that jointly analyze verbal and vocal cues from earnings calls. Qin and Yang \cite{qin-yang-2019-say} proposed a deep learning framework that combines textual content with acoustic features, demonstrating that how executives speak significantly impacts market response. Building upon this foundation, Sawhney et al. \cite{sawhney2020risk} introduced a neural architecture that employs cross-modal attention mechanisms to capture verbal-vocal coherence while also incorporating stock network correlations through graph-based learning. Their approach outperformed previous state-of-the-art methods by augmenting speech features with correlations from text and stock network modalities.

Existing works have evolved from textual sentiment analysis using financial-specific dictionaries \cite{loughran2011liability} to vocal/audio analysis of manager speech patterns \cite{sawhney-etal-2021-empirical}, Graph Neural Networks for text classification, and combined verbal-vocal cue analysis for volatility \cite{sawhney-etal-2020-voltage} and risk \cite{sawhney2020risk} prediction. However, their applicability to emerging markets like India has not been fully explored. Existing studies predominantly focus on US markets, with limited research specifically addressing Indian earnings calls. The distinct characteristics of Indian financial markets—such as regulatory variations, cultural nuances in communication, and unique market dynamics—call for tailored approaches rather than the direct adoption of models designed for Western markets. Additionally, there is a critical need for India-specific datasets and benchmarks to enable thorough evaluation and validation of predictive models in this context.

\section{Problem Statement}

This study addresses the problem of predicting opening stock prices for Indian companies on the day following the release of quarterly earnings results, leveraging multi-modal data (numeric, text transcripts, images from presentations, and tabular data). 





The performance of the proposed framework is evaluated using  Mean Absolute Error (MAE), Root Mean Squared Error (RMSE), and Mean Absolute Percentage Error (MAPE).

\section{Dataset Construction}
The \textbf{MiMIC} (\textbf{M}ult\textbf{i}-\textbf{M}odal \textbf{I}ndian Earnings \textbf{C}alls) dataset was constructed by systematically collecting and processing multi-modal data from earnings calls of Indian companies across different market capitalizations. This comprehensive dataset includes earnings call transcripts, presentation materials, fundamentals, technical indicators, and stock performance metrics to facilitate the analysis of market reactions following corporate disclosures.

\subsection{Company Selection}
We selected all companies representing the Nifty 50 Index, Nifty Midcap 50 index, and Nifty Smallcap 50 index of the Indian stock market as of 3\textsuperscript{rd} November, 2024. For each company, we collected their NSE ticker symbols from their respective company profile pages, which served as unique identifiers throughout our data collection process. We had to eliminate certain companies due to the non-availability of sufficient information. Finally, we were left with 133 companies.

\subsection{Multi-Modal Data Collection}
For each selected company, we gathered the following data components from January 2019 to November 2024:

\begin{itemize}
    \item \textbf{Textual Data:} Earnings call transcripts were collected from Screener.in \footnote{\url{https://www.screener.in/} (accessed on 30\textsuperscript{th} November, 2024)} Text-heavy slides underwent Optical Character Recognition (OCR) to extract textual information.
    
    \item \textbf{Visual Data:} Presentation slides used during earnings calls were collected from the same website and visual elements such as charts, graphs, and images were preserved in their original format for visual analysis.
    
    \item \textbf{Tabular Data:} Financial tables from presentations were extracted separately using image2table \footnote{\url{https://github.com/xavctn/img2table} (accessed on 28\textsuperscript{th} March, 2025)} to maintain their structural integrity, as they often contain critical quantitative information about company performance.

    \item \textbf{Numeric Data:} We incorporated a range of numerical features, encompassing technical and fundamental indicators, macro-economic variables  and market data, into our analysis. A comprehensive set of these variables as presented in \S \ref{sec:appendix-earning-numeric}

\end{itemize}


\subsection{Stock Performance Data}
To establish the relationship between earnings calls and subsequent market reactions, we collected stock price data for each company:
\begin{itemize}
    \item Opening price on the day of earnings call ($d$)
    \item Opening price on the day following \footnote{Note: We are using opening price of the next day and not the opening price of the day of earnings call because most of these calls happen after the market hours \href{https://www.etnownews.com/markets/tcs-infosys-wipro-hcl-tech-q4-results-fy-2025-date-time-dividend-update-quarterly-earnings-schedule-article-151356517}{https://www.etnownews.com/markets/tcs-infosys-wipro-hcl-tech-q4-results-fy-2025-date-time-dividend-update-quarterly-earnings-schedule-article-151356517}} earnings call ($d+1$) 
\end{itemize}

We attempted to collect audio data for earnings calls, but it was unavailable in the majority of cases. The initial dataset underwent a cleaning process to remove instances where both the earnings call transcript and the corresponding presentation slides were unavailable. This resulted in a final dataset of 1,042 instances, derived from 768 transcripts and 833 presentations.

To evaluate the performance of the proposed models, we partitioned the dataset into three distinct subsets based on temporal criteria. Data spanning up to February 7, 2024, was allocated to the training set (80\% of the total data). Data from February 8, 2024, to August 9, 2024, was used for validation (10\%), and data beyond August 10, 2024, was reserved for testing (10\%).

\section{Experiments \& Results}


Our experimental approach progressed through the following stages of feature incorporation:

\begin{enumerate}
\item \textbf{Numeric Features}: We initially utilized only numeric features (N). We trained various machine learning models (like Extreme Random Forest \cite{geurts2006extremely}, Distributed Random Forest (DRF) \cite{h2o-drf},  XGBoost \cite{xgboost}, Gradient Boosting Machine  \cite{gbm}, Feed forward neural network based Deep Learning (\textbf{DL-1}), etc.)  for regression using the AutoML framework of H2O. \footnote{\href{https://docs.h2o.ai/h2o/latest-stable/h2o-docs/automl.html}{https://docs.h2o.ai/h2o/latest-stable/h2o-docs/automl.html} (accessed on 8\textsuperscript{th} April, 2025)} The \textbf{DL-1} model performed the best.

\item \textbf{Text Features}: We expanded our feature set by incorporating textual data (T) from transcripts, presentations, and tables in markdown format. To represent these textual features, we employed the Nomic 1.5 \cite{nussbaum2024nomic} model to extract embeddings (Em). We used matryoshka representation learning to truncate the dimension of embeddings to 128. This was essential as we had only 832 instances to train the regression models. After evaluating multiple H2O AutoML models, the feed-forward neural network (\textbf{DL-2}) demonstrated superior performance. Subsequently, we trained a XGBoost model for binary classification utilizing exclusively text embedding features to predict whether the stock's opening price on day (d+1) would exceed that of day (d). Its F1 score on validation set was 0.675. The predicted probability (P) outputs from this classifier were then incorporated as features in the original regression framework (\textbf{DL-1}), 
thereby creating a cascaded prediction framework. After, training multiple models using H2O AutoML, we obtain best results from a feed forward neural network based model (\textbf{DL-3}).

\item \textbf{Image Features}: We further augmented our dataset with visual information (I).
    We used the Nomic Vision 1.5 model \cite{nussbaum2024nomic} to extract embeddings from images. For instances with multiple images, we applied mean pooling to the image embeddings. Just like the text embeddings, we truncated the dimension of embeddings to 128. Among H2O AutoML models trained on text and image embeddings taken together, the feed-forward neural network (\textbf{DL-4}) yielded optimal results. Following our text-based approach, we similarly trained a DRF model for binary classification using only image embeddings to predict next-day price increases. The F1 score of this classifer was 0.680. The resulting probability estimates were then used as features, 
    in our regression framework (\textbf{DL-3}), 
    extending our cascaded framework from numeric and text to visual data. We followed an identical evaluation process using H2O AutoML, with a feed-forward neural network (\textbf{DL-5}) similarly emerging as the optimal model, mirroring our findings from the text modality.
\end{enumerate}

This stepwise approach allowed us to assess the impact of each feature type on the model's performance. Finally, we evaluated the performance of Llama-4 Maverick \cite{llama4}, a state-of-the-art multi-modal vision language model, under zero-shot conditions (\S \ref{sec:appendix-mimic-prompt}) using raw images and text. The results corresponding to the best performing models for each case are presented in Table \ref{tab:mimic-regression-results}. More details regarding these models and the hyperparameters are provided in the Appendix \S \ref{sec:appendix-mimic-hyper}.

Upon analysis of our experimental results, we observed that direct incorporation of text (T) and image (I) embeddings (Em) as supplementary features to our regression model trained on numeric (N) features resulted in performance degradation. Conversely, when we employed a two-stage approach — first training separate classification models using textual and visual data to generate prediction probabilities (P), then incorporating these probabilities as features in the original regression framework — we achieved significant performance improvements. Our methodological workflow is illustrated in the Appendix \S \ref{sec:appendix-mimic-workflow} (Figure \ref{fig:mimic-workflow}).

Due to constraints in data availability and methodological transparency, a direct comparison with several prior studies was infeasible. Specifically, the models presented in \cite{qin-yang-2019-say}, \cite{sawhney2020risk}, \cite{sawhney-etal-2020-voltage}, and \cite{sawhney-etal-2021-empirical} could not be replicated, as their implementations rely on audio features which were not included in our dataset. Furthermore, the model proposed in 
\cite{medya-www-22} is not open source, preventing a comparative analysis. 

\begin{table}[]
\centering
\caption{Results. Details of the models are mentioned in \S \ref{sec:appendix-mimic-hyper}. Deep Learning (DL), Numeric (N), T (Text), I (Image), Embedding (Em), Predicted Probabilities (P) }
\label{tab:mimic-regression-results}. 
\resizebox{.5\textwidth}{!}{%
\begin{tabular}{llrrr} \hline
\textbf{Model} & \textbf{Modalities} & \textbf{MAE}         & \textbf{RMSE}        & \textbf{MAPE}        \\ \hline
DL-1 & N                & 150.769 & 269.193 & \textbf{0.288} \\
DL-2 & N+ T (Em)         & 228.321 & 348.152 & 0.454          \\
DL-3 & N+ T (P)          & 125.204 & 216.639 & 0.349          \\
DL-4 & N+ T (Em) + I (Em) & 271.350 & 457.369 & 0.965          \\

DL-5           & N+ T (P) + I (P)      & \textbf{104.787}     & \textbf{188.537}     & 0.334                \\
Llama-4        & N + T (Raw) + I (Raw) & \multicolumn{1}{l}{108.417} & \multicolumn{1}{l}{246.196} & \multicolumn{1}{r}{ 5.918} \\ 
\hline
\end{tabular}%
}
\end{table}

\section{Conclusion}
In this study, we have introduced \textbf{MiMIC}, a novel multi-modal dataset, alongside a comprehensive framework for predicting Indian stock price movements following earnings call announcements. Our findings demonstrate that both the textual transcripts and visual presentations from earnings calls significantly influence subsequent stock price behavior, albeit through different mechanisms. The integration of these complementary information sources through our cascaded prediction framework yields superior performance compared to unimodal approaches. Despite recent advances in vision-language models, our experiments reveal that state-of-the-art architectures still face limitations when applied to specialized financial forecasting tasks.

This research lays the groundwork for several promising avenues of future investigation. One key direction involves expanding the current multi-modal framework to incorporate the audio modality. Converting the visual elements like charts and plots into texts, and incorporating them in the model is a potential avenue for additional research. Another area for future exploration is to move beyond next-day price predictions and investigate more granular, intra-call price movements. Developing models capable of predicting stock price fluctuations during the earnings call itself would be of significant practical value to traders and investors.

\bibliography{custom}

\section*{Limitations}
Despite the promising results of our study, several limitations must be acknowledged. First, our analysis is restricted to a sample of 133 companies, which, while representative of the Nifty indices, may not capture the full diversity of the Indian corporate landscape. Expanding this dataset to include a broader range of companies could enhance the generalizability of our findings. 

Second, our methodology only incorporates instances where both stock price data and comprehensive earnings call materials (transcripts and presentations) were available, potentially introducing selection bias by excluding companies with incomplete documentation.

Third, due to computational resource constraints, we employed smaller language models rather than state-of-the-art larger models, which might have limited the depth of linguistic understanding in our analysis. 

Finally, our current approach does not account for variations in speaking styles, audio data characteristics, or presentation formats, which could contain valuable predictive information beyond the textual and visual content analysed. Future research should address these limitations to develop more robust and comprehensive models for predicting stock price movements following corporate earnings calls.

\label{sec:limitations}
\appendix

\section{Appendix}
\label{sec:appendix}
\subsection{Details of Numeric Data}
\label{sec:appendix-earning-numeric}

\subsubsection{Macroeconomic Variables:} Gross Domestic Product (GDP) Growth, Inflation Rate

\subsubsection{Market Data:} NIfty 50 Opening Price, Nifty 50 Closing Price, Nifty 50 Volume

 \subsubsection{Technical Indicators:} Simple Moving Averages (SMA20, SMA50), Relative Strength Index (RSI14)

\subsubsection{Fundamental Indicators:}
A comprehensive set of fundamental variables  was collected for each company. Due to the annual frequency of this data, we utilized the previous year's values for training and prediction.
\textbf{Financial statement items} (Sales, Expenses, Operating Profit, Other Income, Interest Expense, Depreciation, Profit Before Tax, Tax Rate, Net Profit, EPS, Dividend Payout, Equity Capital, Reserves, Borrowings, Other Liabilities, Total Liabilities, Fixed Assets, CWIP, Investments, Other Assets, Total Assets), \\
\textbf{Cash flow items} (Cash from Operating Activities, Cash from Investing Activities, Cash from Financing Activities, Net Cash Flow),\\
\textbf{Additional metrics} (Revenue, Financing Profit, Financing Margin, Deposits, Borrowing)

\subsection{Hyper-parameters}
\label{sec:appendix-mimic-hyper}
The hyper-parameters of the models discussed in this paper, are presented here.
\subsubsection{Text Embedding based classifier}
Model Type: XGBoost \\
Number of trees: 30
\subsubsection{Image Embedding based classifier}
Model Type: Distributed Random Forest\\
Number of trees: 40\\
minimum depth: 13, maximum depth: 20  \\  
minimum leaves: 94, maximum leaves: 115    \\

\subsubsection{Regression Model}
Model Type: Feed-forward based neural network (DL-5)\\
Number of layers: 3\\
Number of hidden units: 20\\
Dropout: 10\\

Hyper-parameters of other models (i.e., DL-1 to DL-4) and other information in detail are provided in the code base \footnote{ 
\url{https://huggingface.co/datasets/sohomghosh/MiMIC\_Multi-Modal\_Indian\_Earnings\_Calls\_Dataset/blob/main/MiMIC\_analysis\_code.ipynb}}.
\subsection{Workflow}
\label{sec:appendix-mimic-workflow}
\begin{figure*}[t]
  \includegraphics[width=\textwidth]{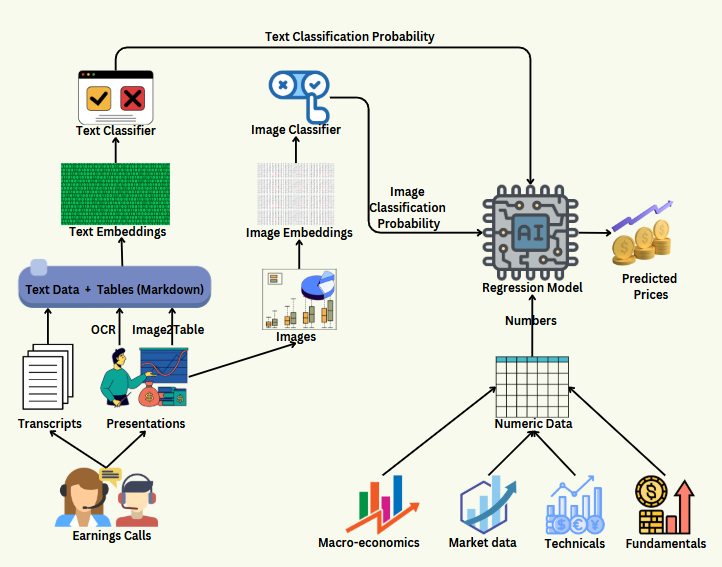}
  \caption{Workflow}
  \label{fig:mimic-workflow}
\end{figure*}

Our methodological workflow is illustrated in Figure \ref{fig:mimic-workflow}.

\subsection{Reproducibility}
The codes and the datasets can be accessed from Hugging Face   
\url{https://huggingface.co/datasets/sohomghosh/MiMIC\_Multi-Modal\_Indian\_Earnings\_Calls\_Dataset/}

\subsection{Prompt}
\label{sec:appendix-mimic-prompt}
You are an expert financial analyst. Using the earnings call transcript, images from the presentation slides, technical indicators, macroeconomic variables, market data, fundamental indicators, and the opening price on the earnings release day, estimate the opening stock price of the company on the day next to the day of the earnings call. Only provide the answer as a real number. No need for any justification. \\ Input Text: \textit{<text along with tables in markdown format>} \\ Input Numeric: \textit{<numeric data along with column names in json format>} \\ Input Images: \textit{<list of input images>}

\end{document}